**Towards a Psychology of Machines: Large Language Models Predict Human Memory**

Markus Huff[1,2] & Elanur Ulakçı[1,2]

[1] Leibniz-Institut für Wissensmedien, Tübingen, Germany

[2] Eberhard Karls Universität Tübingen, Germany

**Author Note**

Correspondence concerning this article should be addressed Markus Huff; Leibniz-Institut für Wissensmedien, Schleichstr. 6, 72072 Tübingen, Germany.

Email: markus.huff@uni-tuebingen.de

**Abstract**

Large language models (LLMs), such as ChatGPT, have shown remarkable abilities in natural language processing, opening new avenues in psychological research. This study explores whether LLMs can predict human memory performance in tasks involving garden-path sentences and contextual information. In the first part, we used ChatGPT to rate the relatedness and memorability of garden-path sentences preceded by either fitting or unfitting contexts. In the second part, human participants read the same sentences, rated their relatedness, and completed a surprise memory test. The results demonstrated that ChatGPT's relatedness ratings closely matched those of the human participants, and its memorability ratings effectively predicted human memory performance. Both LLM and human data revealed that higher relatedness in the unfitting context condition was associated with better memory performance, aligning with probabilistic frameworks of context-dependent learning. These findings suggest that LLMs, despite lacking human-like memory mechanisms, can model aspects of human cognition and serve as valuable tools in psychological research. We propose the field of machine psychology to explore this interplay between human cognition and artificial intelligence, offering a bidirectional approach where LLMs can both benefit from and contribute to our understanding of human cognitive processes.

**Towards a Psychology of Machines: Large Language Models Predict Human Memory**

Transformer-based large language models (LLMs) have revolutionized the area of natural language processing with their exceptional performance, offering profound implications for the science of psychology, given the integral role of language in all subfields (Demszky et al., 2023). Despite their inherent lack of subjective experience possessed by humans, such as thinking or feeling, LLMs are arguably more than mere stochastic replicators of their input statistics. They are designed to produce human-like realistic linguistic outputs, and they are trained on extensive datasets crafted by humans themselves. These attributes have markedly elevated the prominence of LLMs in a field we introduce as *machine psychology*. Although–from a linguistic perspective–LLMs are no models of cognitive processes ("Language Models and Linguistic Theories beyond Words," 2023), studies in this field explore the capabilities of LLMs and the degree to which they align with human behavior and cognitive processes (Aher et al., 2023; Binz & Schulz, 2023b; Buschoff et al., 2023; Dhingra et al., 2023; Dillion et al., 2023; Gilardi et al., 2023; Horton, 2023; Michelmann et al., 2023; Ritter et al., 2017; Tak & Gratch, 2023). Taken together, these studies emphasize a remarkable congruity between human cognition and LLMs, indicating the potential of LLMs to play a meaningful role in psychological research. In this paper, we go beyond the current understanding of LLMs, particularly ChatGPT, by investigating its potential for predicting human memory performance.

**Machine Psychology: Exploring the Intersection of Human and Artificial Cognition**

Following their launch, LLMs have exhibited exceptional capabilities in natural language processing (NLP) and production (Manning et al., 2020). Ranging from the simplest interactions as chatbots to their implementation in various areas such as education (Kasneci et al., 2023), healthcare (Thirunavukarasu et al., 2023), and information search (Stadler et al., 2024), these models have woven themselves into human experiences with their remarkable advantages and the convenience they provide. Beyond their expertise in NLP and practical applications, LLMs reveal significant alignment with human cognitive functions and behaviors in psychological tests where they are evaluated as subjects, despite not being fundamentally developed for this purpose. To effectively utilize LLMs, and to better understand the increasing harmony between them and humans, it is crucial to examine the strengths and limitations of these models. Nonetheless, the lack of transparency in how exactly LLMs function poses a challenge (Schwartz, 2022; Zubiaga, 2024). To address this challenge and to open up these "black boxes", psychology offers a strong framework with its multidisciplinary perspective. In this paper, we propose *machine psychology*, an area dedicated to explore the interplay between human cognition and behavior, and the capabilities of LLMs. In the field of machine psychology, LLMs are perceived as a distinct yet analogous "second mind" that is yet to be uncovered, focusing on their resemblance to human cognitive functions while acknowledging their artificial nature.

This emerging field adopts a bidirectional approach, integrating the unique capabilities of LLMs and the rich and profound knowledge of psychology to deepen the understanding of both artificial and human cognition. On one hand, psychological knowledge and methodologies can be applied to understand LLMs through examining them as participants in assessments and experiments. This approach enables the systematic exploration of their internal processes, behavioral tendencies, and cognitive patterns, as well as the extent to which these models mirror human cognition (Binz & Schulz, 2023b; Mei et al., 2024; Webb et al., 2023). On the other hand, LLMs possess significant potential to advance the study of human cognition. Their distinct information processing mechanisms open new avenues for research and provide the opportunity to use LLMs as proxies/simulations for investigating human cognitive processes, thereby broadening our understanding of the human mind (Binz & Schulz, 2023a; Horton, 2023). Drawing from the concept of *Symbiotic Autonomy* in human-robot interaction studies, which focuses on the reciprocal cooperation between humans and robots (Vanzo et al., 2020), we propose a similar bidirectional approach of machine psychology, where humans and LLMs mutually contribute to advance the understanding of the complexities of both human and artificial cognition. Our focus in this paper is on the latter direction, utilizing LLMs as a proxy to investigate human cognitive processes. For the first time, we apply LLMs to predict human memory performance. We achieve this by conducting a language-based memory task that incorporates garden-path sentences and contextual information. In the following sections, we examine the implications of LLMs' ability to predict the memory performance of humans, despite their absence of a human-like memory mechanism.

**Context-Driven Memory: A Framework for Evaluating LLMs' Predictions**

Our investigation into the potential of LLMs to predict human memory performance is centered on the critical role of context in the formation of memories. Context is important in resolving ambiguity (Szewczyk & Federmeier, 2022), the pervasive and fundamental element of the natural language, which poses a challenge to comprehension (MacGregor et al., 2020). The recognized detrimental impact of ambiguity on language processing is not exclusive to humans, it also constitutes a significant impediment for LLMs (Irwin et al., 2023; Liu et al., 2023). While GPT employs an autoregressive design with the sequential generation process, BERT builds upon this approach by utilizing a bidirectional architecture, considering both preceding and following contexts (Naveed et al., 2023). Despite possessing advanced linguistic processing capabilities, both models encounter difficulties in effectively managing the ambiguity inherent in language, mirroring the challenges present in human cognitive processing during comprehension (Irwin et al., 2023; Liu et al., 2023). This shared challenge of managing ambiguity is particularly evident in the interpretation of complex sentence structures, such as garden-path sentences (Fujita, 2021; Li et al., 2024). Although grammatically correct, these sentences introduce temporary ambiguity which can lead to misinterpretation and hinder comprehension (Ferreira et al., 2001). While garden-path sentences present serious challenges for understanding due to inherent ambiguity, prior context is acknowledged to reduce those complications (Grodner et al., 2005; Kaiser & Trueswell, 2004). The beneficial impact of prior context on comprehending garden-path sentences can be explained through the structure-building framework (Gernsbacher, 1997). This framework suggests that comprehension involves constructing mental representations by linking incoming information to an established structure. In this case, the relevant context sentence provides a foundation for the mental structure, onto which the garden-path sentence can be integrated. Being part of the same mental representation, the context sentence aids in clarifying ambiguity enabling the cohesive processing of both sentences and improving comprehension (Brich et al., 2024). The help of prior context is not only important for facilitating understanding, but also plays a key role for memory formation. Successful comprehension, achieved by extracting the meaning of sentences (Kaup et al., 2024), thereby envisioning the situations described in the text (Gernsbacher & Robertson, 1995; Schütt et al., 2023) lays the groundwork for creating robust mental representations which are more likely to be retained over time (Kintsch et al., 1990). In the case of not being able to comprehend the sentence, readers make use of the surface form of information (Schnotz & Bannert, 2003), known to be forgotten well before the meaning of the text (*verbatim effect;* Poppenk et al., 2008). By aiding understanding, prior context enhances the encoding of garden-path sentences into memory, establishing a critical link between comprehension and retention.

**Experimental Overview and Hypotheses**

Despite lacking a foundation in human cognition, LLMs achieve near-human performance across various tasks (Binz & Schulz, 2023b). This begs the question: can LLMs, if not simply "stochastic parrots" (Digutsch & Kosinski, 2023), reveal insights into the underlying mechanisms of human cognition? Here, we investigate this by harnessing generative AI to predict human memory performance. Specifically, we test if LLMs can predict how well humans remember garden path sentences prefaced by fitting or unfitting contexts. This study probes the potential of LLMs to shed light on human information processing, even while operating on different principles.

**Table 1.** An exemplary representation of the prompts and sentence pairs submitted to ChatGPT.

| Context | Relatedness prompt | Memorability prompt |
|---|---|---|
| Fitting | Read Sentence 1 and Sentence 2 and answer the following question. How related are the two sentences from 1 (not at all) to 10 (highly)?<br><br>Sentence 1: "Bill has chronic alcoholism."<br>Sentence 2: "Because Bill drinks wine is never kept in the house." | Read Sentence 1 and Sentence 2 and answer the following question. How do you rate the memorability of Sentence 2 from 1 (not at all) to 10 (excellent)?<br><br>Sentence 1: "Bill has chronic alcoholism."<br>Sentence 2: "Because Bill drinks wine is never kept in the house." |
| Unfitting | Read Sentence 1 and Sentence 2 and answer the following question. How related are the two sentences from 1 (not at all) to 10 (highly)?<br><br>Sentence 1: "Bill likes to play golf."<br>Sentence 2: "Because Bill drinks wine is never kept in the house." | Read Sentence 1 and Sentence 2 and answer the following question. How do you rate the memorability of Sentence 2 from 1 (not at all) to 10 (excellent)?<br><br>Sentence 1: "Bill likes to play golf."<br>Sentence 2: "Because Bill drinks wine is never kept in the house." |

*Note*: Sentence 2 always represented the garden path sentence.

In the first part of this study, we submitted the garden-path sentences (Sentence 2 in the prompt) with a preceding sentence (Sentence 1 in the prompt) that matched (fitting) or mismatched (unfitting) the context of the garden-path sentence to ChatGPT (Table 1). We collected 100 responses for each prompt (*relatedness* of Sentences 1 and 2, and *memorability* of Sentence 2; including a robustness check with synonyms). In the second part of this study, we used the LLM responses to predict human performance, in which we presented participants with the same material and asked them about the relatedness of the two sentences. After that, we presented participants with a surprise memory test in which we presented only the garden path sentences and measured recognition memory.

We hypothesize that LLMs' and humans' relatedness values are higher in the fitting than the unfitting condition. Further, we hypothesize LLMs' memorability ratings to be higher in the fitting than the unfitting condition. Eventually, we hypothesize that LLM memorability responses would predict participants' memory performance in a surprise memory test.

## Method

We report how we determined our sample size, all data exclusions (if any), all data inclusion/exclusion criteria, whether inclusion/exclusion criteria were established prior to data analysis, all manipulations, and all measures in the study. The experiment was approved by the local ethics committee of the Leibniz-Institut für Wissensmedien (LEK 2023/051).

### Data Sources

*LLM*. We used OpenAI's API to access ChatGPT (OpenAI et al., 2023) (model: GPT-4; June 2023) to collect 100 responses (consisting of the relatedness and the memorability values) for each sentence pair. We set the temperature value to 1 to increase the variability in the answers. This resulted in 4500 independent responses in the fitting and 4500 in the unfitting condition and resulted in a total of 9000 independent responses.

*Participants.* We recruited 100 English-only speaking participants via Prolific. 15 participants indicated that their vision was not normal or not corrected-to-normal during the experiment (i.e., they did not wear lenses or glasses). Thus, the resulting sample consisted of 85 participants (57 female, 27 male, 1 w/o response), mean age was $M = 45.34$ years ($SD = 13.95$)

**Material**

*Garden-path sentences.* We compiled a list of 45 garden-path sentences (e.g., "Because Bill drinks wine is never kept in the house"). For each garden-path sentence, we constructed a sentence matching the context of the garden-path sentence (*fitting* context; e.g., "Bill has chronic alcoholism.") and a sentence not matching its context (*unfitting* context; e.g., "Bill likes to play golf."; for the complete list, see Supplementary Table 1). For the machine data, we used all 45 garden-path sentences; for the human data, we omitted the sentence with ID 8 for counter-balancing reasons. ID 8 was chosen for omission because its sentence structure closely resembles that of ID 45, making its exclusion less critical compared to omitting other sentences.

*Prompts.* We submitted zero-shot prompts to ChatGPT regarding *relatedness* and *memorability*, which we presented before both sets of sentence pairs, one with a fitting context and the other with an unfitting context sentence preceding the garden-path sentence. First, we gave the prompt based on the category. Then, we provided the two sentences separately in an order as "Sentence 1" and "Sentence 2" respectively, by "Sentence 1" being the prior context sentence and "Sentence 2" the garden-path sentence (Table 1). By the relatedness prompt, we requested ChatGPT to rate the relatedness of the sentences by giving a value from 1 (not at all) to 10 (highly). Afterward, we requested a value from ChatGPT to indicate the memorability of Sentence 2 (i.e. garden-path sentence) from 1 (not at all) to 10 (excellent).

*Human experiment*. The experiment was programmed with PsychoPy (Peirce, 2022). All instructions and stimuli appeared in white on a gray background. The stimulus material consisted of sentence pairs, which included a garden-path sentence and its preceding context sentence. Each pair was arranged in a visually specific format, with each sentence starting on a new line, one below the other.

**Procedure and Design Human Data**

In the learning phase, participants read 22 sentence pairs comprised of a prior context sentence and a garden-path sentence. Half of the sentences shown were in the fitting condition, the other half were in the unfitting condition. Participants read pairs of sentences at their own pace and proceeded to the next pair by pressing the spacebar. The response button (i.e., spacebar) was activated three seconds after stimulus onset to ensure that the sentences were not skipped and were read by the participants. After each sentence pair, participants rated the relatedness of the two sentences by clicking on a value on the 10-point rating scale presented to them (1: "not at all" to 10: "highly"). After completing the learning phase of the experiment, participants completed a surprise old/new recognition memory test, including 44 garden-path sentences (22 targets from the learning phase and 22 distractors) without their contexts. Participants indicated whether they remembered the sentence shown on screen from the learning phase by pressing the right arrow key for "yes" and the left arrow key for "no" allowing for the calculation of sensitivity ($d'$) from signal detection theory(Green & Swets, 1966). The study employed a one-factorial design with context (fitting, unfitting) as the within-subjects factor. Four counter-balancing conditions ensured that the garden-path sentences were assigned equally to the conditions (fitting vs. unfitting context, target vs. distractor) across participants. The experiment lasted approximately 15 minutes.

## Results

### Machine data (LLM data)

*Relatedness as a function of context.* We fitted a linear mixed-effect model with context (fitting, unfitting) as fixed effect and sentence ID as random intercept. We submitted the resulting model to a type 2 ANOVA (Fox & Weisberg, 2010). The relatedness of the two sentences is higher in the fitting than the unfitting condition, $\chi^2(1) = 44843.00, p < .001$, constituting a successful manipulation check of the context manipulation (Figure 1A).

*Memorability as a function of context.* Similar to the relatedness analysis, we fitted a linear mixed-effect model with context (fitting, unfitting) as fixed effect and sentence ID as the random intercept. Submitting the resulting model to a type 2 ANOVA showed a significant main effect of context, $\chi^2(1) = 2660.00, p < .001$. In the fitting context condition, the memorability of the garden-path sentence was rated higher than in the unfitting context condition (Figure 1B).

**Figure 1**: Relatedness (A) and memorability (B) measures of the machine data and relatedness (C) and memory (D) of the human data as a function of context.

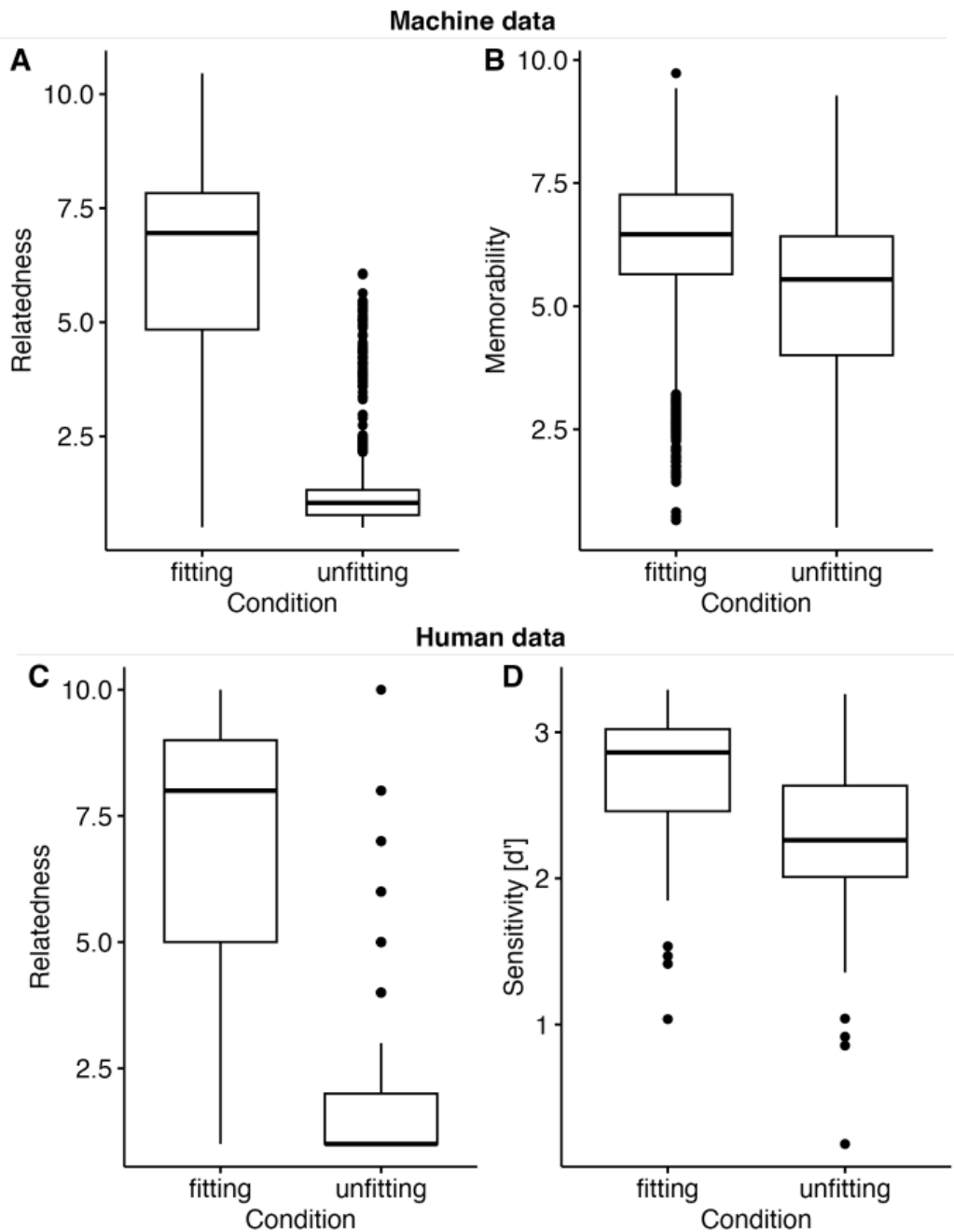

*Robustness check.* Because LLMs are sensitive to small changes in the prompts (Binz & Schulz, 2023b), we repeated the relatedness and the memorability analysis with the following prompts: *How closely are the two sentences linked from 1 (not at all) to 10 (highly)?* and *How do you rate the recognizability of Sentence 2 from 1 (not at all) to 10 (excellent)?* In particular, we changed *related* with *linked* and *memorability* with *recognizability*.

Results resembled the main analysis (see also Supplementary Figure 1). In the fitting condition, the two sentences were judged to be more linked, $M = 6.67$ ($SD = 2.25$), than in the unfitting condition, $M = 1.12$ ($SD = 0.59$), $\chi^2(1) = 43731.60$, $p < .001$. Further, the recognizability values were higher in the fitting, $M = 2.67$ ($SD = 1.34$), than in the unfitting condition, $M = 1.00$ ($SD = 0.02$), $\chi^2(1) = 43731.60$, $p < .001$. We thus conclude that the observed effects are stable.

**Human data (Human experiment)**

*Relatedness as a function of context.* Analysis was similar to the machine data with the exception that we additionally included participant as random intercept. The relatedness of the two sentences is higher in the fitting than the unfitting condition, $\chi^2(1) = 4087.60$, $p < .001$, again constituting a successful manipulation check and replicating the machine data (Figure 1C).

*Recognition memory as a function of context.* To assess participants' memory, we calculated $d'$ from signal detection theory (Green & Swets, 1966), which corrects for response bias. We corrected for perfect performance. In particular, in case of *no false alarms*, we added 0.5 (i.e., a half trial), and in case of all hits, we subtracted 0.5 (i.e., a half trial). A t-test for repeated measures showed a significantly higher recognition performance in the fitting context than in the unfitting context, $t(84) = 5.75$, $p < .001$, Cohen's $d = 0.62$ (Figure 1D).

*Response bias (c).* Participants' responses were more liberal in the fitting ($M = -0.08$, $SD = 0.22$) than in the unfitting condition ($M = 0.13$, $SD = 0.29$), $t(84) = -5.78$, $p < .001$, Cohen's $d = -0.63$.

Our concluding analysis explored a potential mechanism underlying the observed effect, grounded in a probabilistic framework of context-dependent learning and stochastic reasoning. This framework posits that memory performance improves as a function of the probabilistic accumulation and integration of retrieval cues, with context acting as a latent variable that organizes and constrains memory processes (Heald et al., 2023). More precisely, in the fitting condition, where a significant, uniform segment of information spans two sentences, memory performance likely peaks. Thus, relatedness and memory performance should be less related as compared to the unfitting condition, where the relatedness of the two sentences is lower and potentially more heterogeneous. In the latter condition, there is more room for improvement in memory performance. Consequently, in this condition, we expect a significant positive relation, with increases in relatedness of the two sentences leading to noticeable improvements in memory performance. If this is true, we should observe a significant interaction between context (fitting vs. unfitting) and the degree of relatedness, with higher relatedness values, predicting higher memory performance in the unfitting condition but not necessarily in the fitting condition.

To test this framework on the machine data, we fitted a linear mixed effect model with memorability as the dependent variable, context (fitting, unfitting) as a categorical fixed effect, relatedness as a continuous fixed effect, and sentence id as a random intercept (Figure 2A). Submitting this model to a type-2 ANOVA showed a significant interaction of context and relatedness, $\chi^2(1) = 86.37$, $p < .001$. As predicted, in the unfitting condition, higher relatedness values predict higher memory performance. This effect is weaker in the fitting condition. Further, the main effects of condition, $\chi^2(1) = 259.51$, $p < .001$, and relatedness, $\chi^2(1) = 31.23$, $p < .001$, were significant.

**Figure 2**: Memory performance as a function of condition (fitting, unfitting) and relatedness for the machine data (A) and the human data (B).

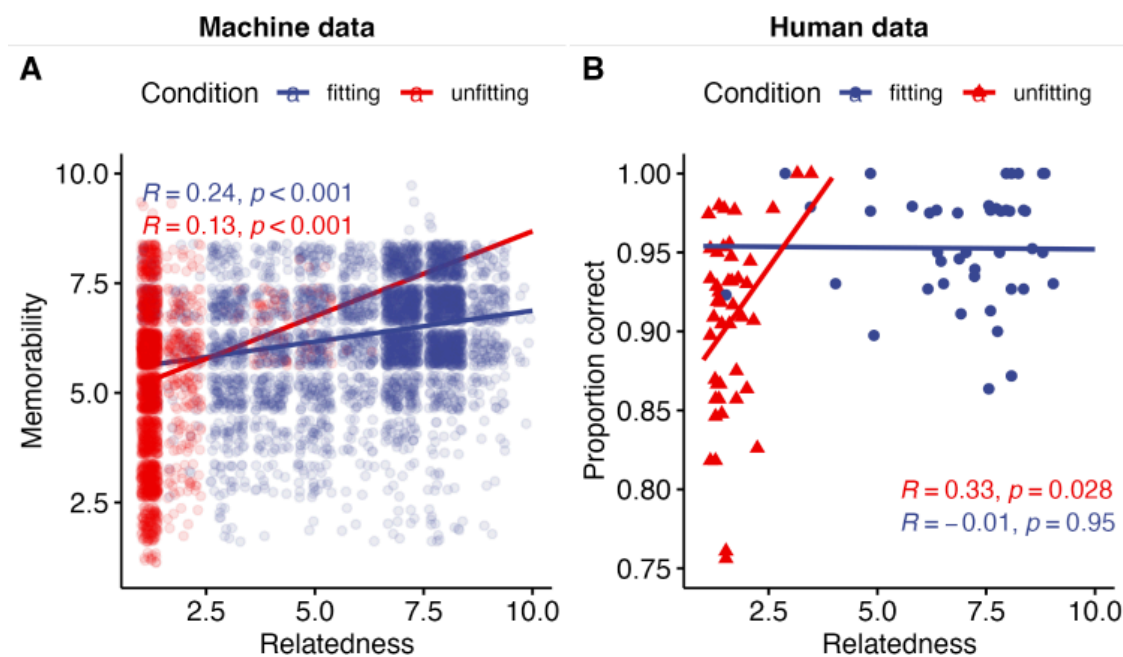

For the analysis of the human data, we aggregated the proportion correct data[1] and the relatedness data at the sentence and context level. We fitted a linear mixed effect model with proportion correct as the dependent variable, context (fitting, unfitting) as a categorical fixed effect, relatedness as a continuous fixed effect, and sentence id as a random intercept (Figure 2B). Submitting this model to a type 2 ANOVA showed a significant interaction of context and relatedness, $\chi^2(1) = 9.39$, $p = .002$. As predicted, in the unfitting condition, higher relatedness values predict higher memory performance, whereas we did not observe such an effect in the fitting condition. Further, the main effects of condition, $\chi^2(1) = 1.13$, $p = .289$, and relatedness, $\chi^2(1) = 1.39$, $p = .239$, were not significant.

---

[1] Note that the nature of this analysis - linking relatedness and memory performance data - required an analysis on the item level. Thus, we used the proportion correct data and not sensitivity (i.e. *d'*).

Consistent with the proposed stochastic reasoning framework, a significant interaction between context and relatedness emerged for machine and human data. While higher relatedness boosted memory in the unfitting condition, this effect was weaker in the fitting condition. These findings across machine and human data emphasize how the link between relatedness and memory hinges on context and available retrieval cues.

## Discussion

We investigated whether generative AI, although lacking a foundation in human cognition, can predict human cognitive performance based on language-based memory tasks. The results showed that the relatedness ratings of ChatGPT closely corresponded with those of the human participants and that the memorability ratings of ChatGPT indeed predicted the memory performance of humans in the surprise memory test. An analysis to check the robustness of the findings with synonyms (*recognizable* and *linked* for *memorable* and *related*) confirmed the results.

The findings from this study provide strong evidence that LLMs, such as ChatGPT, have significant potential to be utilized as experimental tools for the investigation of human cognition in the context of machine psychology. ChatGPT's ability in predicting human memory performance even in the absence of a human-like memory system highlights its potential to represent cognitive patterns observed in humans. Particularly, the significant alignment between the relatedness ratings of human subjects and those of ChatGPT underscores the model's competence in processing and analyzing the linguistic information in a manner similar to humans. Moreover, the accuracy of the model in predicting the performance of humans on the memory test also highlights the potential of LLMs as proxies for investigating the human memory processes. The results also indicate that, even though LLMs are built on a fundamentally different architecture, they display outcomes that parallel human cognitive processes, particularly in tasks that involve the understanding of contextual reasoning. Corroborated through analyses using synonyms, the robustness of our findings strengthens the soundness of this approach and demonstrates the great possibility of LLMs functioning as effective tools in psychological research.

The outcomes of this study emphasize the *bidirectional benefits of machine psychology*: not only LLMs benefit from psychological approaches for deeper evaluation, but researchers also gain a new perspective to examine the complexities of human cognition. We described one potential mechanism based on the interaction of context and relatedness observed in both the machine and human data, which can be effectively explained through a *stochastic reasoning framework* grounded in the principles of context-dependent learning and probabilistic cue integration (Heald et al., 2023). This framework posits that memory performance is influenced by the probabilistic accumulation of retrieval cues during encoding and retrieval processes, as context serves as a latent variable that organizes and constrains memory representations over time. In the fitting context condition, in which sentences share high semantic relatedness, the abundance of overlapping cues enhances encoding strength and facilitates retrieval, resulting in consistently high memory performance (Polyn et al., 2009). Consequently, variations in relatedness within this condition have a minimal impact because the retrieval cues are already robust. Conversely, in the unfitting context condition, which begins with lower relatedness and fewer available cues, any increase in relatedness significantly boosts the availability of retrieval cues, leading to noticeable improvements in memory performance (Jonker et al., 2013). This aligns with the encoding specificity principle, which posits that memory retrieval is most effective when the context at encoding matches the context at retrieval (Tulving & Thomson, 1973). By modeling how variations in contextual relatedness influence the stochastic processing of retrieval cues, this framework provides a theoretical basis for understanding how both humans and LLMs process linguistic structures (Bhatia & Richie, 2024; Binz & Schulz, 2023b). Heald et al. (2023) further highlight that the probabilistic nature of context inference, reliant on dynamically evolving distributions of context-specific cues, contributes significantly to the robustness and adaptability of memory systems. This broader perspective enriches our understanding of how relatedness and contextual variability interact to influence memory processes in both humans and machine learning models. These results provide a foundation for further study of LLMs as proxies and experimental tools in psychology, offering novel possibilities to discover and represent human cognition and behavior.

**Limitations**

Despite our thorough approach to conducting this research, there are a few limitations that should be considered. First, there can be a potential interplay between fluency and context, particularly in terms of relatedness (Oppenheimer, 2008). Participants might have processed sentences in the fitting condition easier, increasing the familiarity for those items, therefore making the fluency heuristic a potential contributor to the memory process during the recognition test as the relatedness of sentences increases.

Second, LLMs are trained based on extensive datasets, which most probably also include the garden-path sentences we used in our study, and this could be a factor influencing their behavior (*performativity problem)* (Horton, 2023). When creating the stimuli, we deliberately chose not to create them entirely independently. Instead, we compiled a diverse set of garden-path sentences with various types of linguistic ambiguities from a wide range of sources within the literature (see Supplementary Table 1). This approach allowed us to expose both humans and GPT to these sentences in a manner that reflects the way LLMs encounter and process information— through diverse and heterogeneous datasets created by a variety of contributors. By gathering input from multiple perspectives, we sought to enhance the representativeness of our stimuli, thereby establishing a realistic and comprehensive foundation for our investigation.

The present findings were derived employing GPT-4 (OpenAI et al., 2023). Although this model was released in 2023, we are confident that, given the robust alignment between human cognitive performance and the model's predictions, the newer models perform at a similar level.

## Conclusions

LLMs are at the forefront of research in the area we introduce as machine psychology, owing to their remarkable language processing capabilities. Our research revealed ChatGPT's ability to make accurate predictions for the performance of human memory despite not possessing it. It carries important potential for utilizing LLMs in studying human cognition.

## Data Availability Statement

Data and analysis scripts (for the statistical programming language R) have been made publicly available via Zenodo and can be accessed at https://doi.org/10.5281/zenodo.10696562.

## Acknowledgment

This study was part of EU's Erasmus scholarship.

## Author Contributions

MH and EU developed the study concept and design. EU designed the stimuli and programmed the human experiment. MH performed data collection with ChatGPT and performed the data analyses. All authors drafted the manuscript. All authors approved the final version of the manuscript for submission.

GENERATIVE ARTIFICIAL INTELLIGENCE PREDICTS HUMAN MEMORY        24(2023). We're Afraid Language Models Aren't Modeling Ambiguity. In H. Bouamor, J. Pino, & K. Bali (Eds.), *Proceedings of the 2023 Conference on Empirical Methods in Natural Language Processing* (pp. 790–807). Association for Computational Linguistics. https://doi.org/10.18653/v1/2023.emnlp-main.51

MacGregor, L. J., Rodd, J. M., Gilbert, R. A., Hauk, O., Sohoglu, E., & Davis, M. H. (2020). The Neural Time Course of Semantic Ambiguity Resolution in Speech Comprehension. *Journal of Cognitive Neuroscience*, *32*(3), 403–425. https://doi.org/10.1162/jocn_a_01493

Manning, C. D., Clark, K., Hewitt, J., Khandelwal, U., & Levy, O. (2020). Emergent linguistic structure in artificial neural networks trained by self-supervision. *Proceedings of the National Academy of Sciences*, *117*(48), 30046–30054. https://doi.org/10.1073/pnas.1907367117

Mei, Q., Xie, Y., Yuan, W., & Jackson, M. O. (2024). A Turing test of whether AI chatbots are behaviorally similar to humans. *Proceedings of the National Academy of Sciences*, *121*(9), e2313925121. https://doi.org/10.1073/pnas.2313925121

Michelmann, S., Hasson, U., & Norman, K. A. (2023). Evidence That Event Boundaries Are Access Points for Memory Retrieval. *Psychological Science*, 09567976221128206. https://doi.org/10.1177/09567976221128206

Naveed, H., Khan, A. U., Qiu, S., Saqib, M., Anwar, S., Usman, M., Akhtar, N., Barnes, N., & Mian, A. (2023). *A Comprehensive Overview of Large Language Models* (Version 10). arXiv. https://doi.org/10.48550/ARXIV.2307.06435

OpenAI, Achiam, J., Adler, S., Agarwal, S., Ahmad, L., Akkaya, I., Aleman, F. L., Almeida, D., Altenschmidt, J., Altman, S., Anadkat, S., Avila, R., Babuschkin, I., Balaji, S., Balcom, V., Baltescu, P., Bao, H., Bavarian, M., Belgum, J., … Zoph, B. (2023). *GPT-4 Technical Report* (No. arXiv:2303.08774). arXiv. https://doi.org/10.48550/arXiv.2303.08774

Oppenheimer, D. M. (2008). The secret life of fluency. *Trends in Cognitive Sciences*, *12*(6), 237–241. https://doi.org/10.1016/j.tics.2008.02.014

Polyn, S. M., Norman, K. A., & Kahana, M. J. (2009). A context maintenance and retrieval model of organizational processes in free recall. *Psychological Review*, *116*(1), 129–156. https://doi.org/10.1037/a0014420

GENERATIVE ARTIFICIAL INTELLIGENCE PREDICTS HUMAN MEMORY    25Poppenk, J., Walia, G., McIntosh, A. R., Joanisse, M. F., Klein, D., & Köhler, S. (2008). Why is the meaning of a sentence better remembered than its form? An fMRI study on the role of novelty‑encoding processes. *Hippocampus*, *18*(9), 909–918. https://doi.org/10.1002/hipo.20453

Ritter, S., Barrett, D. G. T., Santoro, A., & Botvinick, M. M. (2017). Cognitive Psychology for Deep Neural Networks: A Shape Bias Case Study. *Proceedings of the 34th International Conference on Machine Learning*, 2940–2949. https://proceedings.mlr.press/v70/ritter17a.html

Schnotz, W., & Bannert, M. (2003). Construction and interference in learning from multiple representation. *Learning and Instruction*, *13*(2), 141–156. https://doi.org/10.1016/S0959-4752(02)00017-8

Schütt, E., Dudschig, C., Bergen, B. K., & Kaup, B. (2023). Sentence-based mental simulations: Evidence from behavioral experiments using garden-path sentences. *Memory & Cognition*, *51*(4), 952–965. https://doi.org/10.3758/s13421-022-01367-2

Schwartz, M. D. (2022). Should artificial intelligence be interpretable to humans? *Nature Reviews Physics*, *4*(12), 741–742. https://doi.org/10.1038/s42254-022-00538-z

Stadler, M., Bannert, M., & Sailer, M. (2024). Cognitive ease at a cost: LLMs reduce mental effort but compromise depth in student scientific inquiry. *Computers in Human Behavior*, *160*, 108386. https://doi.org/10.1016/j.chb.2024.108386

Szewczyk, J. M., & Federmeier, K. D. (2022). Context-based facilitation of semantic access follows both logarithmic and linear functions of stimulus probability. *Journal of Memory and Language*, *123*, 104311. https://doi.org/10.1016/j.jml.2021.104311

Tak, A. N., & Gratch, J. (2023). Is GPT a Computational Model of Emotion? *2023 11th International Conference on Affective Computing and Intelligent Interaction (ACII)*, 1–8. https://doi.org/10.1109/ACII59096.2023.10388119

Thirunavukarasu, A. J., Ting, D. S. J., Elangovan, K., Gutierrez, L., Tan, T. F., & Ting, D. S. W. (2023). Large language models in medicine. *Nature Medicine*, *29*(8), 1930–1940. https://doi.org/10.1038/s41591-023-02448-8

**Supporting Information Text**

Table **S1**: List of the used garden-path, fitting context, and unfitting context sentences.

| ID | Fitting Context | Unfitting Context | Garden-Path Sentence | Reference |
|---|---|---|---|---|
| 1 | Bill has chronic alcoholism. | Bill likes to play golf. | Because Bill drinks wine is never kept in the house. | Ferreira, F., & Henderson, J. M. (1991). Recovery from misanalyses of garden-path sentences. *Journal of Memory and Language*, *30* (6), 725-745. |
| 2 | Elderly people are in control. | The birds flew to the south. | The old man the boat. | Di Sciullo, A. M. (2005). UG and External Systems. In *Linguistik aktuell*. https://doi.org/10.1075/la.75 |
| 3 | The soldiers won't have problems about accommodation. | A terrible accident happened down the street. | The complex houses married and single soldiers and their families. | Petrie, H.; Darzentas, J.; Walsh, T. (2016). Universal Design 2016: Learning from the Past, Designing for the Future: Proceedings of the 3rd International Conference on Universal Design (UD 2016), York, United Kingdom, August 21 – 24, 2016. IOS Press. p. 463. ISBN 9781614996842. |
| 4 | He suddenly leaves without telling anyone. | The climate change becomes worse every day. | The man who hunts ducks out on weekends. | Slabakova, R. (2016). *Second Language Acquisition*. Oxford University Press. |
| 5 | She wasn't agreeing about kids being loud. | A huge wildfire started in the north. | I convinced her children are noisy. | Slabakova, R. (2016). *Second Language Acquisition*. Oxford University Press. |
| 6 | His job was to adjust the pianos. | The dodo bird became extinct years ago. | The man who whistles tunes pianos. | Slabakova, R. (2016). *Second Language Acquisition*. Oxford University Press. |
| 7 | The oil people consume is unhealthy. | The school president elections will be held today. | Fat people eat accumulates. | Slabakova, R. (2016). *Second Language Acquisition*. Oxford University Press. |

| ID | Fitting Context | Unfitting Context | Garden-Path Sentence | Reference |
|---|---|---|---|---|
| 8 | Tom was doing his chores in the kitchen. | She was feeling very tired today. | While Tom was washing the dishes fell on the floor. | DCODR Wordsmithing tools. (n.d.). https://wordsmithingtools.com/list-of-garden-path-sentences |
| 9 | The men of the village go deer hunting on weekends. | The new videos of an UFO are all over social media. | When the men hunt the birds typically scatter. | Ferreira, F., & Henderson, J. M. (1991). Recovery from misanalyses of garden-path sentences. *Journal of Memory and Language*, *30* (6), 725-745. |
| 10 | Attacking a dog is a very dangerous thing. | He bought the favorite game of his friend as a birthday present. | When the boys strike the dog kills. | Ferreira, F., & Henderson, J. M. (1991). Recovery from misanalyses of garden-path sentences. *Journal of Memory and Language*, *30* (6), 725-745. |
| 11 | Martians won the war against human beings. | His favorite goldfish died. | After the Martians invaded the town was evacuated. | Ferreira, F., & Henderson, J. M. (1991). Recovery from misanalyses of garden-path sentences. *Journal of Memory and Language*, *30* (6), 725-745. |
| 12 | The boy was itching all over. | He said he hated his new haircut. | While the boy scratched the dog yawned loudly. | Ferreira, F., & Henderson, J. M. (1991). Recovery from misanalyses of garden-path sentences. *Journal of Memory and Language*, *30* (6), 725-745. |
| 13 | The concert of the orchestra was great at the beginning. | France is the 13th country I have been to. | After the musician played the piano was wheeled off of the stage. | Goldstein, E. B. (2014). Cognitive psychology: Connecting mind, research and everyday experience. Cengage Learning. |
| 14 | The man was in the forest expecting to hunt a bird. | He bought the painting for 45 million Euros. | While the man hunted the deer ran into the woods. | Christianson, K., Hollingworth, A., Halliwell, J. F., & Ferreira, F. (2001). Thematic roles assigned along the garden path linger. *Cognitive psychology, 42* (4), 368-407. |
| 15 | The lawyer wanted to take a look at the proofs for the homicide. | I became a yoga instructor after 7 years of work. | The evidence examined by the lawyer turned out to be unreliable. | Ferreira, F., & Clifton Jr, C. (1986). The independence of syntactic processing. *Journal* |

| ID | Fitting Context | Unfitting Context | Garden-Path Sentence | Reference |
|---|---|---|---|---|
|  |  |  |  | *of memory and language*, 25 (3), 348-368. |
| 16 | Fred always wastes his meals. | She fell down the stairs and broke her arm. | When Fred eats food gets thrown. | Slabakova, R. (2016). *Second Language Acquisition.* Oxford University Press. |
| 17 | The author commented about a novel on the forum website. | He stayed hungry all day due to forgetting his wallet. | The author wrote the novel was likely to be a best-seller. | *Towards Natural Language Processing: A Well-Formed Substring Table Approach to Understanding Garden Path Sentence.* (2010, November 1). IEEE Conference Publication | IEEE Xplore. https://ieeexplore.ieee.org/abstract/document/5659102 |
| 18 | The child was attacked by a dog. | She was studying for her finals very hard. | Mary gave the child the dog bit a bandaid. | Slabakova, R. (2016). *Second Language Acquisition.* Oxford University Press. |
| 19 | There are very bad-tempered people around. | He became the new editor of the journal. | The sour drink from the ocean. | *"The horse raced past the barn fell": A guide to garden path sentences.* (n.d.). https://effectiviology.com/avoid-garden-path-sentences-in-your-writing/ |
| 20 | She helped a lot on this project. | The party got cancelled due to weather conditions. | Without her contributions would be impossible. | *"The horse raced past the barn fell": A guide to garden path sentences.* (n.d.). https://effectiviology.com/avoid-garden-path-sentences-in-your-writing/ |
| 21 | The dog never liked John. | She was so angry because her partner forgot their 7th anniversary of their wedding day. | Wherever John walks the dog chases him. | *"The horse raced past the barn fell": A guide to garden path sentences.* (n.d.). https://effectiviology.com/avoid-garden-path-sentences-in-your-writing/ |
| 22 | He never gets tired of running. | He forgot to close the window now the room is all wet due to rain. | Because he always jogs a mile seems a short distance to him. | *"The horse raced past the barn fell": A guide to garden path sentences.* (n.d.). https://effectiviology.com/avoid-garden-path-sentences-in-your-writing/ |

| ID | Fitting Context | Unfitting Context | Garden-Path Sentence | Reference |
|---|---|---|---|---|
| 23 | I was searching for an article related to my project. | Everyone was shocked about the death of the popular singer. | While I was surfing the internet went down. | *"The horse raced past the barn fell": A guide to garden path sentences*. (n.d.). https://effectiviology.com/avoid-garden-path-sentences-in-your-writing/ |
| 24 | He was having his lunch. | The animals in the farm were dying because of a serious disease. | While the man was eating the pizza was still being reheated in the oven. | *"The horse raced past the barn fell": A guide to garden path sentences*. (n.d.). https://effectiviology.com/avoid-garden-path-sentences-in-your-writing/ |
| 25 | This neighborhood is not safe. | A drunk driver hit a tree yesterday night. | Until the police arrest the criminals control the street. | Slabakova, R. (2016). *Second Language Acquisition*. Oxford University Press. |
| 26 | The team players and the coach gave a little break from the workout. | She got jealous when she learned she will have a new sister. | The coach smiled at the player tossed the frisbee. | Tabor, W., Galantucci, B., & Richardson, D. C. (2004). Effects of merely local syntactic coherence on sentence processing. Journal of Memory and Language, 50 (4), 355–370. https://doi.org/10.1016/j.jml.2004.01.001 |
| 27 | Elderly people always closely track young people. | He was very upset because the tickets for the concert of his favorite band was sold out. | The old dog the footsteps of the young. | Smyth, M. M. (1994). *Cognition in action*. Psychology Press. |
| 28 | Amber's mother prepared a nice dinner for thanksgiving. | She couldn't decide which shoes to wear for her outfit. | While Amber hunted the turkey was on the table. | Schütt, E., Dudschig, C., Bergen, B. K., & Kaup, B. (2023). Sentence-based mental simulations: Evidence from behavioral experiments using garden-path sentences. *Memory & Cognition*, *51* (4), 952-965. |
| 29 | Zoe finally had some alone time. | The formula 1 race in Monaco was very exciting. | As Zoe bathed the baby slept in the bed. | Schütt, E., Dudschig, C., Bergen, B. K., & Kaup, B. (2023). Sentence-based mental simulations: Evidence from behavioral experiments using garden-path sentences. |

| ID | Fitting Context | Unfitting Context | Garden-Path Sentence | Reference |
|---|---|---|---|---|
| | | | | *Memory & Cognition*, *51* (4), 952-965. |
| 30 | Ryan won the tough race. | The raise in prices made the citizens very angry. | While Ryan won the car was in poor condition. | Schütt, E., Dudschig, C., Bergen, B. K., & Kaup, B. (2023). Sentence-based mental simulations: Evidence from behavioral experiments using garden-path sentences. *Memory & Cognition*, *51* (4), 952-965. |
| 31 | Samuel went to the new and nice restaurant. | The newly graduated architect got the very prestiged prize. | While Samuel ordered the fish swam upstream. | Schütt, E., Dudschig, C., Bergen, B. K., & Kaup, B. (2023). Sentence-based mental simulations: Evidence from behavioral experiments using garden-path sentences. *Memory & Cognition*, *51* (4), 952-965. |
| 32 | Mary was having her dinner in the evening. | She realized her bike got stolen. | As Mary ate the egg was in the fridge. | Schütt, E., Dudschig, C., Bergen, B. K., & Kaup, B. (2023). Sentence-based mental simulations: Evidence from behavioral experiments using garden-path sentences. *Memory & Cognition*, *51* (4), 952-965. |
| 33 | Edward was working on a new artwork. | She prepared nothing to serve to her guests. | While Edward painted the house was afire. | Schütt, E., Dudschig, C., Bergen, B. K., & Kaup, B. (2023). Sentence-based mental simulations: Evidence from behavioral experiments using garden-path sentences. *Memory & Cognition*, *51* (4), 952-965. |
| 34 | Eve went for a walk around the neighborhood. | I tried bungee jumping for the very first time yesterday. | As Eve walked the dog lay on the ground. | Schütt, E., Dudschig, C., Bergen, B. K., & Kaup, B. (2023). Sentence-based mental simulations: Evidence from behavioral experiments using garden-path sentences. *Memory & Cognition*, *51* (4), 952-965. |

| ID | Fitting Context | Unfitting Context | Garden-Path Sentence | Reference |
|---|---|---|---|---|
| 35 | Miranda was preparing the cocktail. | Because she was late for class, her teacher didn't let her in. | While Miranda stirred the coffee was roasted. | Schütt, E., Dudschig, C., Bergen, B. K., & Kaup, B. (2023). Sentence-based mental simulations: Evidence from behavioral experiments using garden-path sentences. *Memory & Cognition*, *51* (4), 952-965. |
| 36 | Anna was trying on her new clothes. | She got rescued 9 hours later of her ski accident. | While Anna dressed the baby spit up on the bed. | Ferreira, F., Christianson, K., & Hollingworth, A. (2001). Misinterpretations of garden-path sentences: Implications for models of sentence processing and reanalysis. *Journal of psycholinguistic research*, *30*, 3-20. |
| 37 | The chef came up with a delicious recipe. | The ministry warned people about not going out due to adverse weather condition. | The chef declared to the reporters he baked the public would be impressed. | Waters, G. S., & Caplan, D. (1996). Processing resource capacity and the comprehension of garden path sentences. *Memory & Cognition*, *24* (3), 342-355. |
| 38 | The suspect confessed something to his lawyer during court. | He couldn't watch the new episode of the series. | The defendant confided to the lawyer he admired the judge was his brother. | Waters, G. S., & Caplan, D. (1996). Processing resource capacity and the comprehension of garden path sentences. *Memory & Cognition*, *24* (3), 342-355. |
| 39 | The man dropped by his uncle. | He said he hated the food at the restaurant. | While the man was visiting the children who were surprisingly pleasant and funny played outside. | Van Gompel, R. P., Pickering, M. J., Pearson, J., & Jacob, G. (2006). The activation of inappropriate analyses in garden-path sentences: Evidence from structural priming. *Journal of Memory and Language*, *55* (3), 335-362. |
| 40 | The doctor was having his daily visit in the morning. | The best part of the balkan tour was seeing Macedonia. | When the doctor was visiting the patient had a heart attack. | Van Gompel, R. P., Pickering, M. J., Pearson, J., & Jacob, G. (2006). The activation of inappropriate analyses in garden-path sentences: Evidence from structural priming. *Journal of Memory* |

| ID | Fitting Context | Unfitting Context | Garden-Path Sentence | Reference |
|---|---|---|---|---|
|  |  |  |  | *and Language*, *55* (3), 335-362. |
| 41 | We wouldn't achieve anything without her. | She insisted on not paying for the pie she didn't like. | Without her efforts would be in vain. | Luoa, R. (2017). Studies on Garden Path Phenomenon in English. *Sociology*, *7* (7), 371-375. |
| 42 | The tension of people in the tribunes was increasing. | He got very angry in the airport when he saw his luggage exploded. | The referees warned the spectators would probably get too rowdy. | Ning, L. H., & Shih, C. (2012). Prosodic effects on garden-path sentences. In *Speech Prosody 2012*. |
| 43 | The student was reading some facts about books. | The anti-bullying program results show that it is not an effective preventive intervention. | The student read a book is good for the mind. | TASK 1: 'Garden path' sentences. (n.d.). Retrieved February 29, 2024, from https://www.math.uni.wroc.pl/~msliw/lingw/mar5zad1.pdf |
| 44 | The plans of the executives were suggested to the employees for voting. | For tomorrow we plan to go to the beach. | The management plans to cut vacation days are rejected. | Jacqueline. (2016, April 6). *Garden Path Sentences – Crazy Sentences that Do Have Meaning*. English With a Smile. https://englishwithasmile.org/2015/01/31/garden-path-sentences-crazy-sentences-that-do-have-meaning/ |
| 45 | New plans were suggested in the parliament by the potency. | I went on a hike today. | The government plans to raise taxes were defeated. | *"The horse raced past the barn fell": A guide to garden path sentences*. (n.d.-b). https://effectiviology.com/avoid-garden-path-sentences-in-your-writing/ |

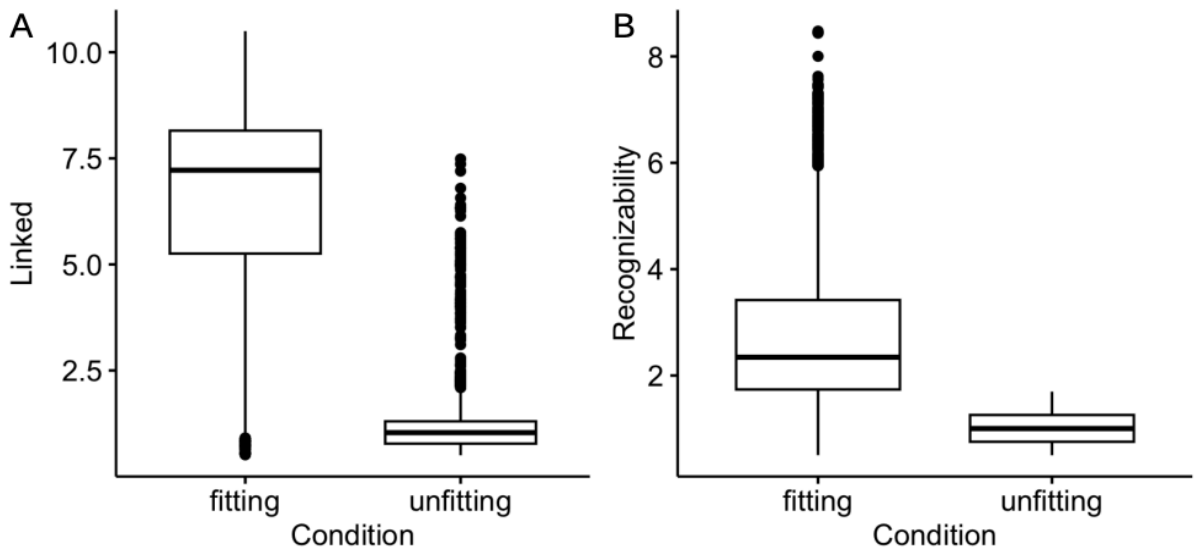

**Fig. S1**: Robustness checks with the following prompts: *How closely are the two sentences linked from 1 (not at all) to 10 (highly)?* and *How do you rate the recognizability of Sentence 2 from 1 (not at all) to 10 (excellent)?*